\pdfoutput=1

\documentclass[11pt]{article}

\usepackage{acl}

\usepackage{times}
\usepackage{latexsym}

\usepackage[T1]{fontenc}

\usepackage[utf8]{inputenc}

\usepackage{microtype}

\usepackage{multirow,booktabs,graphicx,float}
\title{More Discriminative Sentence Embeddings via Semantic Graph Smoothing}

\author{Chakib Fettal$^{1,2}$\and Lazhar Labiod$^1$ \and Mohamed Nadif$^1$\\
  $^1$Centre Borelli UMR 9010, Université Paris Cité, 75006 Paris, France \\
  $^2$Informatique Caisse des Dépôts et Consignations, 75013 Paris, France\\
  \texttt{\{firstname.lastname\}@u-paris.fr} \\
}

\usepackage{amsmath,amsfonts}
\usepackage{flushend}

\newcommand{\mD}{\mathbf{D}}
\newcommand{\mS}{\mathbf{S}}

\newcommand{\mL}{\mathbf{L}}
\newcommand{\mA}{\mathbf{A}}

\newcommand{\mX}{\mathbf{X}}
\newcommand{\mI}{\mathbf{I}}
\newcommand{\mH}{\mathbf{H}}
\newcommand{\vf}{\mathbf{f}}

\newcommand{\vx}{\mathbf{x}}

\newcommand{\better}[1]{\cellcolor{blue!12.5}{#1}}
\usepackage{colortbl}

\begin{document}
\maketitle
\begin{abstract}
This paper explores an empirical approach to learn more discriminantive sentence representations in an unsupervised fashion. Leveraging semantic graph smoothing, we enhance sentence embeddings obtained from pretrained models to improve results for the text clustering and classification tasks. Our method, validated on eight benchmarks, demonstrates consistent improvements, showcasing the potential of semantic graph smoothing in improving sentence embeddings for the supervised and unsupervised document categorization tasks.
\end{abstract}

\section{Introduction}
Text categorization, also known as document categorization, is a natural language processing (NLP) task that involves arranging texts into coherent groups based on their content. It has many applications such as spam detection \cite{jindal2007review}, sentiment analysis \cite{melville2009sentiment}, content recommendation \cite{pazzani2007content}, etc. There are two main approaches to text categorization: classification (supervised learning) and clustering (unsupervised learning). In text classification, the process involves training a model using a labeled dataset, where each document is associated with a specific category. The model learns patterns and relationships between the text features and the corresponding categories during the training phase. Text clustering, however, aims to group similar documents together without prior knowledge of their categories. Unlike text classification, clustering does not require labeled data. Instead, it focuses on finding inherent patterns and similarities in the text data to create clusters.

In the field of NLP, pretrained models have attained state-of-the-art performances in a variety of tasks \cite{devlin-etal-2019-bert,liu2019roberta,reimers2019sentence}, one of which is text classification. In spite of that, text clustering using such models did not garner significant attention. To this day most text clustering techniques use the representations of texts generated by some pretrained model such as Sentence-BERT \cite{reimers2019sentence} and often use classical clustering approaches such as k-means to obtain a partition of the texts. This is done without any fine-tuning due to the unsupervised nature of the clustering problem. 

Recently, graph filtering has appeared as an efficient and effective technique for learning representations for attributed network nodes. The effectiveness of this technique has made it a backbone for popular deep learning architectures for graphs such as the graph convolutional network (GCN) \cite{kipf2016semi}. Simplified versions of this deep architecture have been proposed wherein the learning of large sets of weights has been deemed unnecessary. Their representation learning scheme works similar to Laplacian smoothing and, by extension, graph filtering. We can give as examples of these simplified techniques the simple graph convolution (SGC) \cite{wu2019simplifying}, and the simple spectral graph convolution (S²GC) \cite{zhu2020simple}. Some researchers used GCNs for the task of text classification. \citet{yao2019graph} proposed TextGCN which is GCN with a custom adjacency matrix built from word PMI and the TF-IDF of the documents with the attributes being word count vectors. \citet{lin2021bertgcn} proposed BertGCN which is similar to TextGCN  with the difference that they use BERT representation for the GCN and combine their training losses. The issue is that these approaches are not suitable for learning unsupervised representations since labels are needed. This is a significant limitation towards their use in unsupervised tasks. Recently some graph-based unsupervised approaches were proposed to deal with text data represented using document-term matrices \cite{fettal2022subspace,fettal2023boosting}.

In this paper, we propose to use the concept of graph smoothing/filtering, which is the main component accredited with the success of GCNs \cite{defferrard2016convolutional,kipf2016semi,li2018deeper}, to semantically "fine-tune" the representations obtained via sentence embedding models to help traditional clustering and classification algorithms better distinguish between semantically different texts and group together texts which have similar meanings, all in an unsupervised manner. To do this, we build a graph with respect to the text which describes the semantic similarity between the different documents based on the popular cosine similarity measure. Our approach yields almost systematic improvement when using filtering on the textual representations as opposed to using them without filtering in both facets of document categorization: classification and clustering. Experiments on eight popular benchmark datasets support these observations.

The code for the experiments is available at \footnote{\url{https://github.com/chakib401/smoothing_sentence_embeddings}}.


\section{Background: Graph Filtering and Smoothing}
Graph Signal Processing \cite{shuman2013emerging,ortega2018graph} provides a framework to analyze and process signals defined on graphs, by extending traditional signal processing concepts and tools to the graph domain. This allows for the representation and manipulation of signals in a way that is tailored to the specific structure of the graph.
In what follows we refer to matrices in boldface uppercase and vectors in boldface lowercase.

\paragraph{Graph Signals} Graph signals are mappings from the set of vertices to the real numbers. A graph signal for a given graph $\mathcal{G}$ can be represented using vector $\vf=[f(v_1),\ldots,f(v_n)]^\top$ such that $f:\mathcal{V}\to \mathbb{R}$ is a real-valued function on the vertex set. The smoothness of a signal $\vf$ over graph $\mathcal{G}$ can be characterized using the Laplacian quadratic form associated with Laplacian $\mL$:
\begin{equation}
    \vf^\top\mL\vf=\frac{1}{2}\sum_{i,j}\ a_{ij}(\vf_i-\vf_j)^2.
    \label{eq:smooth}
\end{equation}
These signals can be high dimensional and can represent many kinds of data. In our case, signals will represent text embeddings. 
\paragraph{Graph Filters} Smoother graph signals can be obtained by minimizing the quantity described in Formula (\ref{eq:smooth}). That is the goal of graph filters and the filtering is generally done from a spectral perspective. A specific class of filters that additionally has an intuitive interpretation from a vertex perspective is that of the polynomial filters. When the filter is a $P$-th order polynomial of the form $\hat{h}(\mL)=\sum_{m=0}^{p}\theta_m\mL^m$, the filtered signal at vertex $i$, is a linear combination of the components of the input signal at vertices within a $P$-hop local neighborhood of vertex $i$:
\begin{equation}
    \vf^{\text{out}}_i=\alpha_{ii}\vf^{\text{in}}_i+\sum_{j\in N(i,p)}\alpha_{ij}\vf^{\text{in}}_j
\end{equation}
where $N(i,p)$ is the $P$-th order neighborhood of vertex $i$. It is possible to then make the correspondence with a polynomial filter (from a spectral perspective) as follows: 
\begin{equation}
    \alpha_{ij}=\sum_{m=d_{\mathcal{G}}(i,j)}^{p}\theta_m(\mL^m)_{ij}
\end{equation}
where $d_{\mathcal{G}}$ is the shortest distance between node $i$ and $j$. Several polynomial filters have been proposed in the literature such as the ones associated with Simple Graph Convolution (SGC) \cite{wu2019simplifying}, simple spectral Graph Convolution (S²GC) \cite{zhu2020simple}, approximate personalized propagation of neural predictions (APPNP) \cite{gasteiger2018predict} and Decoupled Graph Convolution (DGC) \cite{wang2021dissecting}. 

\section{Proposed Methodology: Smoothing Sentence Embeddings}

In this paper, we theorize that smoothing sentence embeddings with a semantic similarity graph can help supervised and unsupervised categorization models better differentiate between the similar and dissimilar documents, leading to performance gains. A common choice for quantifying semantic similarity of text is the cosine similarity; {given two sentence embedding vectors $\vx_i, \vx_j \in \mathbb{R}^d$ we have
$$
    \cos(\vx_i, \vx_j)=\frac{\vx_i^\top \vx_j}{\Vert \vx_i \Vert\Vert \vx_j \Vert}.
$$
We build a $k$-nearest neighbors connectivity graph which we denote $\mathcal{G}$ based on this similarity measure i.e. a graph for which each node has exactly $k$ neighbors and whose edge weights are all equal to one. We characterize the graph $\mathcal{G}$ using its adjacency matrix $\mA$, we denote its Laplacian as $\mL$. Given the adjacency matrix, a standard trick to obtain better node representations consists in adding a self-loop \begin{equation}\hat{\mA}=\mA+\lambda\mI\end{equation} where $\lambda$ is a hyperparameter controlling the number of self-loops. As such in what follows we consider the symmetrically normalized version of $\hat{\mA}$, that is 
\begin{equation}
\mS=\hat{\mD}^{-1/2}\hat{\mA}\hat{\mD}^{-1/2}.
\end{equation}
Now given a node embedding matrix $\mX$ and the previous semantic similarity graph. We consider four polynomial graph filters whose propagation rules we describe in Table \ref{tab:rules}.
\begin{table}[!h]
    \caption{The propagation rules associated with the different polynomial filters. $\mH^{(0)}$ is the $\mX$. $P$ is the propagation order. $\alpha$ and $T$ are filter-specific hyperparameters.}
    \centering
    \begin{tabular}{@{}ll@{}}
        \toprule
        Filter & Propagation Rule\\\midrule
        $F_{\text{SGC}}$ &  $\mH^{(p+1)} \gets \mS\mH^{(p)}$ \\
        $F_{\text{S²GC}}$& $\mH^{(p+1)} \gets \mH^{(p)} + \mS\mH^{(p)}$ \\
        $F_{\text{APPNP}}$ & $\mH^{(p+1)} \gets (1-\alpha)\mS\mH^{(p)} + \alpha \mH^{(0)}$ \\
        $F_{\text{DGC}}$ &  $\mH^{(p+1)} \gets (1-\frac{T}{P})\mH^{(p)} + \frac{T}{P}\mS\mH^{(p)}$\\\bottomrule
    \end{tabular}
    \label{tab:rules}
\end{table}

\section{Experiments}
In this section we evaluate our semantically smoothed representations obtained through four filters on two tasks, clustering and classification, with respect to the original representations obtained from SentenceBERT \cite{reimers2019sentence} as well as two large language models baselines: BERT and RoBERTa.

\subsection{Datasets and Metrics}
We use eight benchmark datasets of varying sizes and number of clusters, and we report their summary statistics in Table \ref{tab:datasets}. 
For the metrics, in the supervised context, we use the F1 score as the quality metric while in the unsupervised context we use the adjusted rand index (ARI) \cite{hubert1985comparing} and the adjusted mutual information (AMI) \cite{vinh2009information}.
\begin{table}[!h]
\caption{Summary statistics of the datasets. Balance refers to the ratio of the most frequent class over the least frequent class. Length refers to the average sentence length in the corpus.}
\label{tab:datasets}
\centering
\resizebox{\columnwidth}{!}{
\begin{tabular}{@{}lcccc@{}}
    \toprule
    Dataset & Docs & Classes & Balance & Length\\\midrule
    20News & 18,846 & 20 &1.6&221\\
    DBpedia & 12,000 & 14 &1.1&46\\
    AGNews & 8,000 & 4 &1.1& 31\\
    BBCNews & 2,225 & 5 &1.3&384\\
    Classic3 & 3,891 & 3 &1.4&152\\
    Classic4 & 7,095 & 4 &3.9&107\\
    R8 & 7,674 & 8 & 76.9&65\\
    Ohsumed & 7,400 & 23 & 61.8&135\\\bottomrule
\end{tabular}
}
\end{table}
\begin{table*}[t]
    \centering
    \caption{Clustering results in terms of AMI and ARI on the eight datasets. The best results are highlighted in bold. If our best performing variant outperforms the best comparative method in a statistically significant matter (t-test at a confidence level of 95\%), we highlight it in blue.}
    \label{tab:clustering_results_1}
    \resizebox{\linewidth}{!}{
    \begin{tabular}{@{}lcccccccc@{}}
    \toprule
    & \multicolumn{2}{c}{20News} & \multicolumn{2}{c}{AGNews} & \multicolumn{2}{c}{BBCNews} & \multicolumn{2}{c}{Classic3} \\
    \cmidrule(lr){2-3}\cmidrule(lr){4-5}\cmidrule(lr){6-7}\cmidrule(lr){8-9}
     & AMI & ARI & AMI & ARI & AMI & ARI & AMI & ARI \\\midrule
    ${\text{ENS}}_{\text{BERT-base}}$ & 37.5 \scriptsize{±2.5} & 15.3 \scriptsize{±1.7} & 54.1 \scriptsize{±3.6} & 51.4 \scriptsize{±5.8} & 81.0 \scriptsize{±5.5} & 80.0 \scriptsize{±8.5} & 98.6 \scriptsize{±0.1} & 99.4 \scriptsize{±0.0} \\
    ${\text{ENS}}_{\text{BERT-large}}$ & 46.1 \scriptsize{±0.7} & 21.4 \scriptsize{±0.6} & 58.5 \scriptsize{±2.8} & 58.2 \scriptsize{±5.9} & 86.0 \scriptsize{±3.5} & 86.5 \scriptsize{±6.3} & 98.4 \scriptsize{±0.2} & 99.3 \scriptsize{±0.1} \\
    ${\text{ENS}}_{\text{RoBERTa-base}}$ & 37.5 \scriptsize{±1.4} & 15.9 \scriptsize{±1.8} & 55.9 \scriptsize{±4.1} & 52.1 \scriptsize{±4.1} & 80.0 \scriptsize{±5.3} & 77.2 \scriptsize{±9.4} & 98.4 \scriptsize{±0.1} & 99.3 \scriptsize{±0.1} \\
    ${\text{ENS}}_{\text{RoBERTa-large}}$ & 48.0 \scriptsize{±0.8} & 23.2 \scriptsize{±1.2} & 56.7 \scriptsize{±4.6} & 52.8 \scriptsize{±5.1} & 85.8 \scriptsize{±3.8} & 85.1 \scriptsize{±7.2} & 98.7 \scriptsize{±0.1} & 99.4 \scriptsize{±0.1} \\
    {SBERT+kM} & 62.9 \scriptsize{±0.3} & 47.4 \scriptsize{±1.0} & 57.9 \scriptsize{±0.1} & 60.5 \scriptsize{±0.1} & 90.8 \scriptsize{±0.2} & 93.0 \scriptsize{±0.1} & 96.0 \scriptsize{±0.1} & 97.6 \scriptsize{±0.1} \\\midrule
    {SB+$F_{\text{SGC}}$+kM} & {65.4 \scriptsize{±0.4}} & {49.1 \scriptsize{±1.1}} & \better{\textbf{60.6 \scriptsize{±0.1}}} & {62.4 \scriptsize{±0.3}} & 90.6 \scriptsize{±0.1} & 92.9 \scriptsize{±0.1} & {98.8 \scriptsize{±0.0}} & {99.5 \scriptsize{±0.0}} \\
    {SB+$F_{\text{S²GC}}$+kM} & 64.9 \scriptsize{±0.4} & 49.0 \scriptsize{±1.1} & 60.1 \scriptsize{±0.2} & 62.2 \scriptsize{±0.2} & \textbf{90.9 \scriptsize{±0.1}} & \better{\textbf{93.1 \scriptsize{±0.1}}} & 98.3 \scriptsize{±0.0} & 99.2 \scriptsize{±0.0} \\
    {SB+$F_{\text{APPNP}}$+kM} & {65.4 \scriptsize{±0.4}} & \better{\textbf{49.8 \scriptsize{±1.2}}} & \better{\textbf{60.6 \scriptsize{±0.0}}} & \better{\textbf{62.5 \scriptsize{±0.0}}} & 90.6 \scriptsize{±0.1} & 92.9 \scriptsize{±0.1} & 98.5 \scriptsize{±0.0} & 99.3 \scriptsize{±0.0} \\
    {SB+$F_{\text{DGC}}$+kM} & \better{\textbf{65.6 \scriptsize{±0.7}}} & 48.8 \scriptsize{±1.0} & 60.5 \scriptsize{±1.5} & 60.5 \scriptsize{±2.2} & 90.2 \scriptsize{±0.1} & 92.5 \scriptsize{±0.1} & \better{\textbf{99.1 \scriptsize{±0.0}}} & \better{\textbf{99.6 \scriptsize{±0.0}}} \\
    \end{tabular}
}
\centering
\resizebox{\linewidth}{!}{
\begin{tabular}{@{}lcccccccc@{}}
\midrule
\midrule
 & \multicolumn{2}{c}{Classic4} & \multicolumn{2}{c}{DBpedia} & \multicolumn{2}{c}{Ohsumed} & \multicolumn{2}{c}{R8} \\
 \cmidrule(lr){2-3}\cmidrule(lr){4-5}\cmidrule(lr){6-7}\cmidrule(lr){8-9}
 & AMI & ARI & AMI & ARI & AMI & ARI & AMI & ARI \\
\midrule
${\text{ENS}}_{\text{BERT-base}}$ & 71.4 \scriptsize{±3.5} & 49.0 \scriptsize{±4.0} & 73.4 \scriptsize{±2.5} & 51.0 \scriptsize{±4.0} & 15.2 \scriptsize{±1.0} & 9.1 \scriptsize{±1.2} & 35.3 \scriptsize{±2.0} & 22.7 \scriptsize{±2.4} \\
${\text{ENS}}_{\text{BERT-large}}$ & 73.0 \scriptsize{±1.8} & 51.1 \scriptsize{±3.2} & 72.4 \scriptsize{±2.1} & 47.2 \scriptsize{±4.2} & 16.1 \scriptsize{±0.9} & 9.3 \scriptsize{±0.7} & 35.7 \scriptsize{±3.5} & 22.8 \scriptsize{±3.1} \\
${\text{ENS}}_{\text{RoBERTa-base}}$ & 72.1 \scriptsize{±4.7} & 51.0 \scriptsize{±4.1} & 74.2 \scriptsize{±2.6} & 52.5 \scriptsize{±4.7} & 17.5 \scriptsize{±0.7} & 11.4 \scriptsize{±0.8} & 25.6 \scriptsize{±1.0} & 13.6 \scriptsize{±1.2} \\
${\text{ENS}}_{\text{RoBERTa-large}}$ & 74.1 \scriptsize{±3.5} & 52.5 \scriptsize{±3.9} & 72.5 \scriptsize{±2.5} & 49.0 \scriptsize{±4.4} & 19.4 \scriptsize{±0.7} & 12.7 \scriptsize{±0.7} & 42.4 \scriptsize{±5.6} & 32.9 \scriptsize{±9.2} \\
{SBERT+kM} & 84.5 \scriptsize{±0.1} & 86.2 \scriptsize{±0.1} & 86.0 \scriptsize{±1.4} & 80.0 \scriptsize{±3.1} & 39.3 \scriptsize{±0.7} & 23.5 \scriptsize{±1.2} & 63.1 \scriptsize{±1.8} & 45.5 \scriptsize{±3.7} \\\midrule
{SB+$F_{\text{SGC}}$+kM} & 85.8 \scriptsize{±2.8} & 85.6 \scriptsize{±7.4} & 85.6 \scriptsize{±1.0} & 78.5 \scriptsize{±2.7} & \better{\textbf{41.8 \scriptsize{±0.5}}} & \better{\textbf{25.2 \scriptsize{±1.0}}} & \better{\textbf{65.6 \scriptsize{±0.5}}} & {49.0 \scriptsize{±0.6}} \\
{SB+$F_{\text{S²GC}}$+kM} & 86.0 \scriptsize{±0.0} & {86.9 \scriptsize{±0.0}} & \textbf{86.6 \scriptsize{±1.2}} & \textbf{80.4 \scriptsize{±2.8}} & 41.0 \scriptsize{±0.8} & 24.5 \scriptsize{±1.5} & 64.8 \scriptsize{±1.1} & 47.8 \scriptsize{±0.7} \\
{SB+$F_{\text{APPNP}}$+kM} & {86.2 \scriptsize{±0.0}} & {87.0 \scriptsize{±0.0}} & 85.8 \scriptsize{±1.0} & 78.9 \scriptsize{±2.7} & 41.6 \scriptsize{±0.7} & 24.9 \scriptsize{±1.5} & 65.1 \scriptsize{±1.6} & 48.5 \scriptsize{±1.0} \\
{SB+$F_{\text{DGC}}$+kM} & \better{\textbf{86.9 \scriptsize{±0.0}}} & \better{\textbf{87.7 \scriptsize{±0.0}}} & 85.4\scriptsize{±1.0} & 78.4 \scriptsize{±2.2} & \better{\textbf{41.8 \scriptsize{±0.7}}} & 24.8\scriptsize{±1.7} & \better{\textbf{65.6 \scriptsize{±0.5}}} & \better{\textbf{49.3 \scriptsize{±0.4}}} \\
\bottomrule
\end{tabular}
}
\end{table*}
\begin{table*}[!t]
 \centering
 \caption{Classification results in terms of F1 score on the eight data sets.}
 \label{tab:classification}
\resizebox{\linewidth}{!}{
\begin{tabular}{@{}lcccccccc@{}}
\toprule
 & 20News & R8 & AGNews & BBCNews & Classic3 & Classic4 & DBpedia & Ohsumed \\
\midrule
BERT$_\text{base}$ & 80.7 & 89.94 & \textbf{89.78} & 95.51 & \textbf{100.0} & \textbf{98.58} & \underline{97.84} & 56.48 \\
RoBERTa$_\text{base}$ & 85.48 & 89.42 & 88.06 & 96.73 & 99.16 & 96.47 & \textbf{98.22} & 58.11 \\
SBERT+LR & 83.35 & 90.22 & 86.25 & \underline{98.62} & 99.61 & 98.19 & 97.33 & 62.87 \\\midrule
{SB+$F_{\text{APPNP}}$+LR} & \textbf{87.54} & \underline{90.9} & 87.9 & \textbf{99.06} & \underline{99.75} & 98.36 & 97.14 & \textbf{67.6} \\
{SB+$F_{\text{DGC}}$+LR} & 87.11 & 90.08 & 87.59 & 98.19 & 99.61 & \underline{98.52} & 97.38 & 67.09 \\
{SB+$F_{\text{S²GC}}$+LR} & \underline{87.36} & \textbf{91.19} & \underline{88.33} & \underline{98.62} & \underline{99.75} & 98.19 & 97.26 & \underline{67.42} \\
{SB+$F_{\text{SGC}}$+LR} & 87.26 & 89.22 & 88.05 & \textbf{99.06} & 99.61 & 98.32 & 97.01 & 67.05 \\
\bottomrule
\end{tabular}
}
\end{table*}
\subsection{Experimental Settings}
For the classification task, we use a random stratified 64\%-16\%-20\% train-val-test split. We also tune the hyperparameters $k$ of the $k$-nn graph, order of propagation $P$, the parameter $\lambda$ and the filter specfic parameters $\alpha$ and $T$. For the clustering task, we use $k=10$ for the $k$-nn graph, set $P=2$ as the propagation order, $\lambda=1$, $\alpha=0.1$ and $T=5$. We report the averages of the metrics as well as their standard deviations over 10 runs (for the classification task, we omit standard deviation due to them being insignificant).
\subsection{Experimental Results}

\paragraph{Clustering Results}
We compare the results of the $k$-means algorithm (kM) applied on Sentence-BERT (we refer to it as SBERT or SB) embeddings with and without the different filtering operations. {Note that instead of using kM we can use any other clustering algorithms including variants of kM such as $k$-means$++$ \cite{arthur2007k} and entropy kM \cite{chakraborty2020entropy}.} In addition to this, we add a baseline which uses an ensemble technique \cite{ait2021leverage} on the layer outputs of the word embedding of BERT and RoBERTa, this method improves over considering a single layer or taking the mean. We report the clustering results in Table~\ref{tab:clustering_results_1}. The filtering operation systematically leads to better results on the benchmark with respect to the filterless clustering scheme on all datasets we have used. These increases are statistically significant in most cases. It also significantly beats the ensemble approach on most datasets.

\paragraph{Classification Results}
Similar to the clustering setting, we compare results from a Logistic Regression (LR) applied on the original sentence embeddings with and without the filtering operation we introduced. We also use fine-tuned BERT and RoBERTa (2 epochs) as baselines; we use the base versions due to computational restrictions. We report the results in Table~\ref{tab:classification}. We see that this operation leads to better performances on the classification task on the majority of the datasets with respect to the filterless Sentence-BERT but this performance increase is not as pronounced as for the clustering task. We also see that the representations we learn lead to competitive results with respect to BERT and RoBERTa despite Sentence-BERT not being suited to classification. 

\begin{figure}
    \centering
    \includegraphics[width=1.\columnwidth]{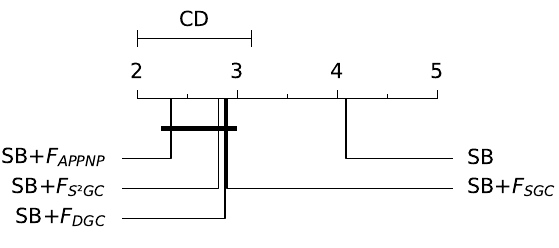}
    \caption{Bonferroni-Dunn average rank test at a confidence level of 95\%.}
    \label{fig:holm}
\end{figure}
\paragraph{Statistical Significance Testing} Using the Bonferroni-Dunn post-hoc mean rank test \cite{demvsar2006statistical}, we analyze the average ranks of the clustering and classification over the Sentence-BERT representations with and without filtering in terms of AMI and ARI, for the clustering task, as well as the F1 score for the classification task on the eight datasets. Figure~\ref{fig:holm} shows that the clustering and classification results when using the proposed semantically smoothed representations are statistically similar and that they all outperform the Sentence-BERT variant with no filtering in a statistically significant manner at a confidence level of 95\%.
\section{Conclusion}
We proposed a simple {yet effective} empirical approach that consists in using similarity graphs in an unsupervised manner to smooth sentence embeddings obtained from pretrained models in a semantically aware manner. The systematic improvements in performance on both clustering and classification tasks on several benchmark datasets of different scales and balance underscore the effectiveness of using semantic graph smoothing to improve sentence representations. 

\section{Limitations}
The main limitation of our approach is the additional computational complexity entailed by creating the $k$-nn graph from the data, performing the smoothing. Add to that, the hyperparameter tuning that is necessary for the classification task. While this increase is in no way prohibitive even for large datasets, a performance-speed compromise is to be considered.


\bibliography{anthology,custom}
\bibliographystyle{acl_natbib}




\end{document}